\documentclass[12pt]{article}

\usepackage{scicite}
\usepackage{float}

\usepackage{times}

\newenvironment{sciabstract}{%
\begin{quote} \bf}
{\end{quote}}

\usepackage{graphicx}
\usepackage{mwe}
\usepackage{sidecap}
\usepackage{amsfonts}
\usepackage{amsthm}
\usepackage{amssymb}
\usepackage{mathtools}
\usepackage{xcolor}
\usepackage{hyperref}
\usepackage{caption}
\usepackage{siunitx}
\usepackage{soul}

\title{Deep learning at the edge enables real-time streaming ptychographic imaging}

\author
{Anakha V Babu,$^{1,\dagger}$ Tao Zhou,$^{1,\dagger}$ Saugat Kandel,$^{1}$ Tekin Bicer,$^{1}$ Zhengchun Liu,$^{1}$ \\ William Judge,$^{2}$ Daniel J. Ching,$^{1}$ Yi Jiang,$^{1}$ Sinisa Veseli,$^{1}$ Steven Henke,$^{1}$ \\ Ryan Chard,$^{1}$ Yudong Yao,$^{1}$ Ekaterina Sirazitdinova,$^{3}$  Geetika Gupta,$^{3}$ \\ Martin V. Holt,$^{1}$  Ian T. Foster,$^{1}$ Antonino Miceli,$^{1,\ast}$ Mathew J. Cherukara $^{1,\ast}$\\
\\
\small{$^{1}$Argonne National Laboratory, 9700 S Cass Ave, Lemont, IL, USA}\\
\small{$^{2}$Formerly at Department of Chemistry, University of Illinois, Chicago, IL, USA}\\
\small{$^{3}$NVIDIA Corporation, Santa Clara, CA, USA}\\
\\
\small{$^\ast$To whom correspondence should be addressed: mcherukara@anl.gov, amiceli@anl.gov}
\\
\small{$^\dagger$Equal contributions}
}

\date{}

\begin{document} 


\baselineskip24pt

\section*{Government License}

The submitted manuscript has been created by UChicago Argonne, LLC,
Operator of Argonne National Laboratory (“Argonne”). Argonne, a
U.S. Department of Energy Office of Science laboratory, is operated
under Contract No. DE-AC02-06CH11357. The U.S. Government retains for
itself, and others acting on its behalf, a paid-up nonexclusive,
irrevocable worldwide license in said article to reproduce, prepare
derivative works, distribute copies to the public, and perform
publicly and display publicly, by or on behalf of the Government.  The
Department of Energy will provide public access to these results of
federally sponsored research in accordance with the DOE Public Access
Plan. http://energy.gov/downloads/doe-public-access-plan.

\newpage


\maketitle 

\begin{sciabstract}
Coherent microscopy techniques provide an unparalleled multi-scale view of materials across scientific and technological fields, from structural materials to quantum devices, from integrated circuits to biological cells. Driven by the construction of brighter sources and high-rate detectors, coherent X-ray microscopy methods like ptychography are poised to revolutionize nanoscale materials characterization. However, these advancements are accompanied by significant increase in data and compute needs, which precludes real time imaging with conventional approaches. Here, we demonstrate a workflow that leverages artificial intelligence at the edge and high-performance computing to enable real-time inversion on X-ray ptychography data streamed directly from a detector at up to \SI{2}{\kilo \hertz}. The proposed AI-enabled workflow eliminates the oversampling constraints, allowing low dose imaging using orders of magnitude less data than required by traditional methods.   

\end{sciabstract}

\section{Introduction}
Ptychography is a high-resolution coherent imaging technique that is widely used in X-ray, optical, and electron microscopy. In particular, X-ray ptychography has the unique potential for non-destructive nanoscale imaging of centimeter-sized objects~\cite{Jiang:2021jw, Du:jo5064} with little sample preparation, and has provided unprecedented insight into countless material and biological specimens. Examples include high resolution imaging of integrated circuits~\cite{PSI-chip} and biological samples~\cite{Deng2015}, as well as strain imaging of nanowires~\cite{Dzhigaev2017,Hill2018} and quantum devices~\cite{Hruszkewycz2017,Shi2019}. Analogously, by leveraging the improved resolution provided by ptychography, electron microscopists have achieved a record breaking deep sub-angstrom resolution~\cite{Jiang:2018jna}. 

Ptychographic imaging is performed by scanning a coherent beam across the sample with a certain degree of spatial overlap, and recording the resulting far-field diffraction patterns. Subsequently, the image is recovered by computationally inverting these measured patterns. The inversion (or image reconstruction) of ptychographic data provides a solution to the phase problem, where only the amplitude information about the sample, and not its phase, is retained in the measured intensities. Currently, real-time imaging with ptychography has not been achieved due to the long computation time of iterative phase retrieval methods. These methods also require a large degree of overlap between adjacent measurement areas to converge, which limits the sample volume that can be imaged in a given amount of time and causes extra damage in beam-sensitive specimens.    

Furthermore, modern ptychography instruments come with a drastically increased data rate, which presents new and prohibitive computational challenges. For example, one raster scan across an area of $\SI{1}{\milli \meter} \times  \SI{1}{\milli \meter}$ with \SI{100}{\nano \meter} step size corresponds to $10^{8}$ points which, with a moderate size detector of one million pixels with 16-bit dynamical range, yields 200 TB of raw data. This volume of data can be acquired in less than 24 hours at a fourth-generation synchrotron source, thus presenting a data rate of \SI{16}{Gbps} and requiring $\sim$PFLOPs of \textit{on-demand} computation to perform phase retrieval \cite{APS_comp_strategy}. To address this challenge, scientists have increasingly turned to deep learning methods for data analysis. An emerging strategy is to replace conventional analysis techniques with much faster surrogate deep learning models~\cite{tomogan,PtychoNN,BraggNN,yao2022autophasenn}. Deep convolutional neural networks have been widely explored for coherent imaging and have been shown to outperform conventional compute-intensive iterative algorithms used for phase retrieval in terms of speed---and increasingly also in reconstruction quality, especially under low-light or other sparse conditions \cite{PtychoNN, DNN_ptychography, Low-photon-DL, Henry_chan_DL, Zhou2021}.

In this work, using X-ray ptychography as a representative technique, we demonstrate an artificial intelligence (AI)-enabled workflow that can keep up with the current maximum detector frame rate of \SI{2}{\kilo \hertz} and provide accurate image reconstructions in real time. We use high performance computing (HPC) resources for online AI training and  a low-cost, palm-sized embedded GPU system at the beamline for the live inference at the edge. This workflow provides a simple and scalable solution to the ever-growing data rate with state-of-the-art ptychography, and can be easily extended to work with other techniques at light sources and advanced electron microscopes.

\section*{Results}

\subsection*{Real-time streaming ptychography imaging workflow}

\begin{figure}[H]
\centering
{\includegraphics[width=1\linewidth]{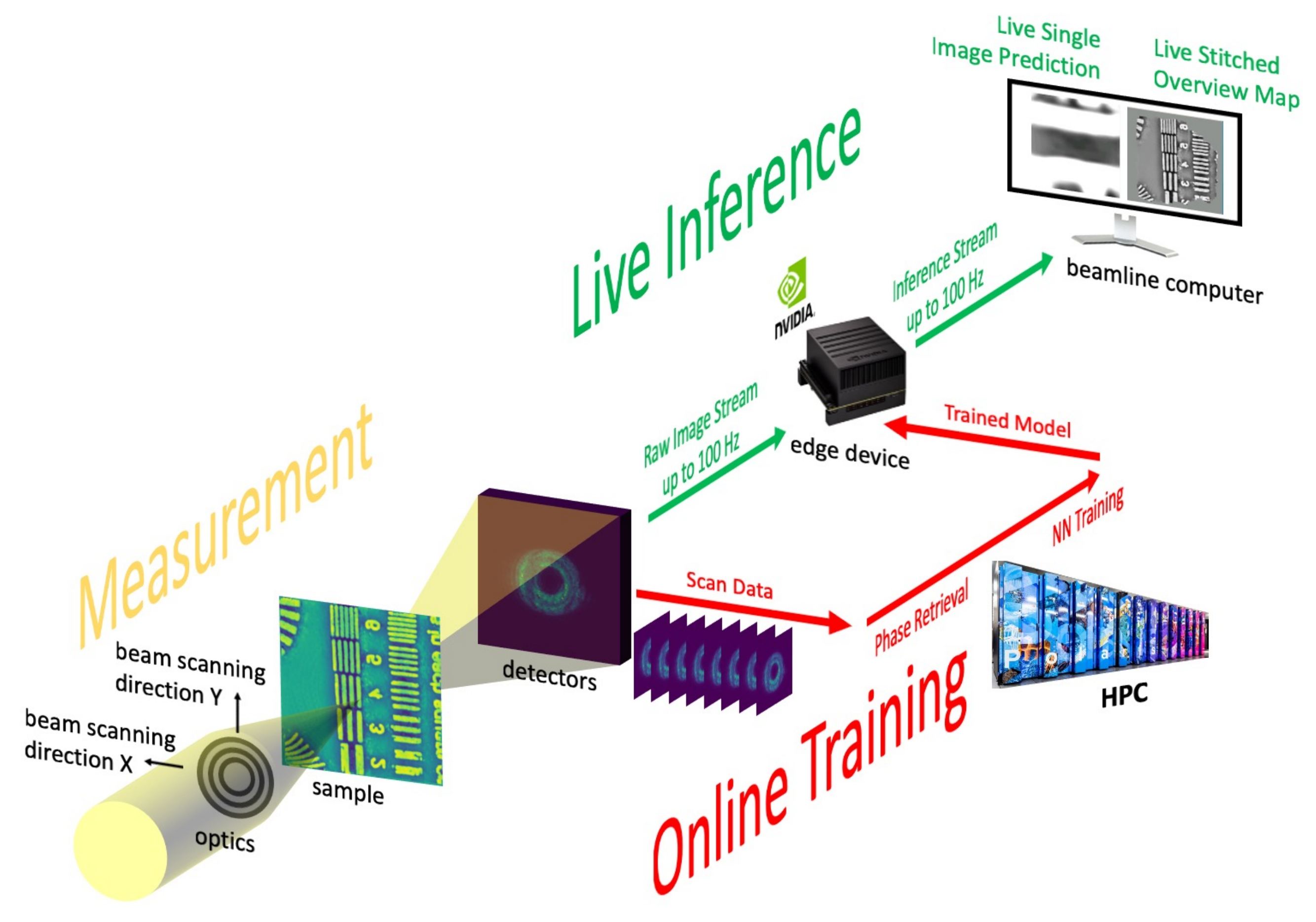}}
\caption{Illustration of AI-enabled workflow for real-time streaming ptychography imaging. An animated version of the sketch can be found here: \url{https://danielzt12.github.io/latest_news/2022/07/11/AI-enabled-on-the-fly-phase-retrieval.html}\cite{video_ref}.}
\label{fig:overall-flow}
\end{figure}

The overall workflow for real-time streaming ptychographic imaging, shown in Figure \ref{fig:overall-flow}, consists of three concurrent components: 1) measurement, 2) online training, and 3) live inference. Diffraction patterns are captured at the detector downstream of the sample while scanning a focused X-ray beam in a spiral pattern. At the end of each scan, the resulting diffraction data are sent to the HPC where real-space amplitude and phase are reconstructed using iterative algorithm ~\cite{ePIE}. Iterative phase retrieval is typically too slow to permit real-time streaming imaging and HPC resources are typically not available on-demand. Instead, the reconstructed real-space images are cropped and paired with the corresponding diffraction patterns to provide labeled data for continued training of a neural net~\cite{PtychoNN}. The diffraction patterns are also streamed concurrently to the edge device at the beamline over the local area network by using a codec-based structured data protocol~\cite{PVA}. Updated with the latest trained model, the edge device infers on individual diffraction patterns, and streams the results back to the beamline computer, providing the users with a stitched sample images in real-time. The workflow is entirely automated, as documented in a video recording provided in the supplementary materials. 

\subsection*{Inference accuracy at the edge}

\color{black}

\begin{figure}[H]
\centerline
{\includegraphics[width=0.95\textwidth]{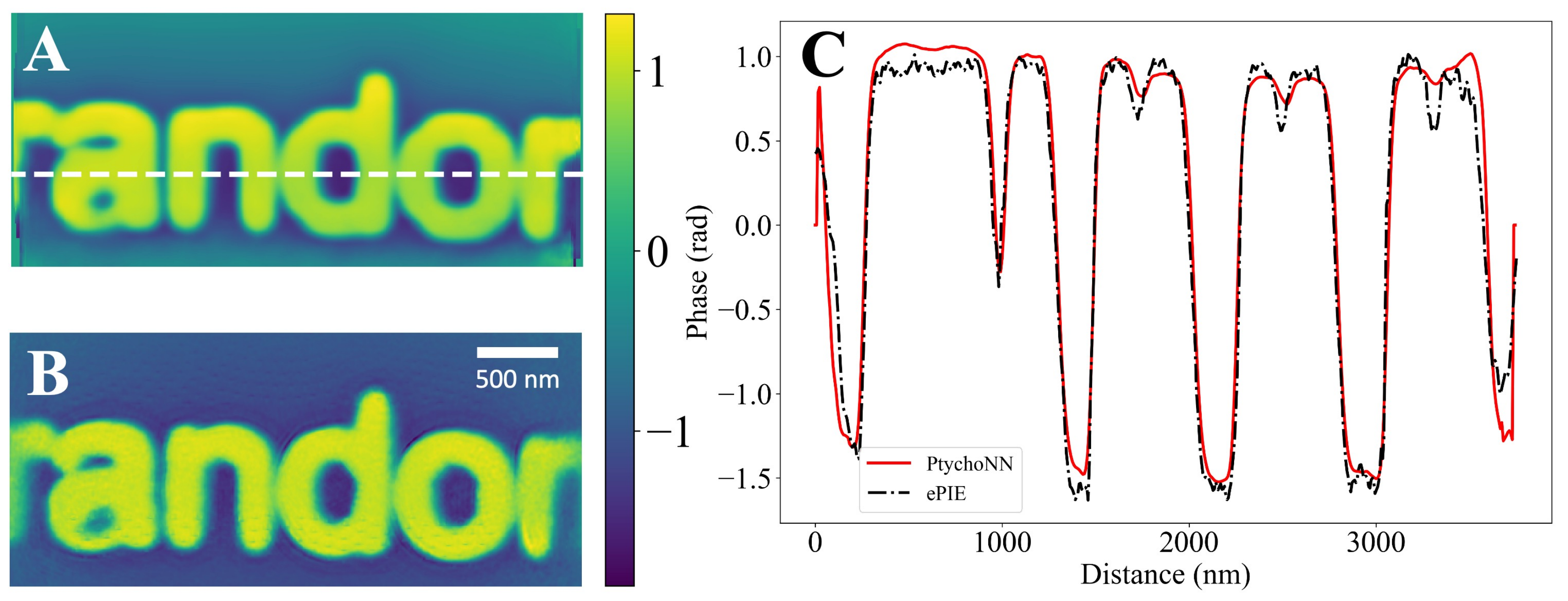}}
\caption{Inference accuracy of the workflow.
\textbf{A} shows the cumulative phase inferred by PtychoNN.  The result is obtained by stitching together individual inferences of the entire scan as described in \cite{PtychoNN}. \textbf{B} shows the corresponding ePIE resconstruction, and \textbf{C} shows the line profile comparison of the phase obtained with PtychoNN and ePIE. The position of the line cut is indicated by the white dashed line in \textbf{A}.}
\label{fig:scan-795}
\end{figure}

We first consider the ability of the workflow to reproduce results from iterative phase retrieval under the same experimental conditions. Figure \ref{fig:scan-795} compares the PtychoNN-inferred and ePIE-reconstructed phases for a test scan with high spatial overlap. A spiral scan was used with a step-size of \SI{50}{\nano \meter} and a beamsize of \SI{750}{\nano \meter}, corresponding to an overlap ratio of 0.88. Here the overlap ratio is defined as $1-S/B$, where \textit{S} is the step-size and \textit{B} is the beamsize. Comparison of the line profiles in Figure \ref{fig:scan-795}C shows that, once trained, PtychoNN inference results are almost identical to those obtained with ePIE. Additionally, we note that the scanned area has features (alphabetical letters) that were not present in the training dataset. Supplementary figure \ref{fig:SF_trainingset} shows an example of the data included in the online training of PtychoNN which consists of entirely random patterns, although their refractive indices remain the same as the training and test data merely correspond to different regions of the same sample etched into diverse features.

\color{black}

\subsection*{Low-dose ptychographic imaging through sparse-sampling}

\begin{figure}[H]
\centerline
{\includegraphics[width=\textwidth,trim=5mm 0 0 0,clip]{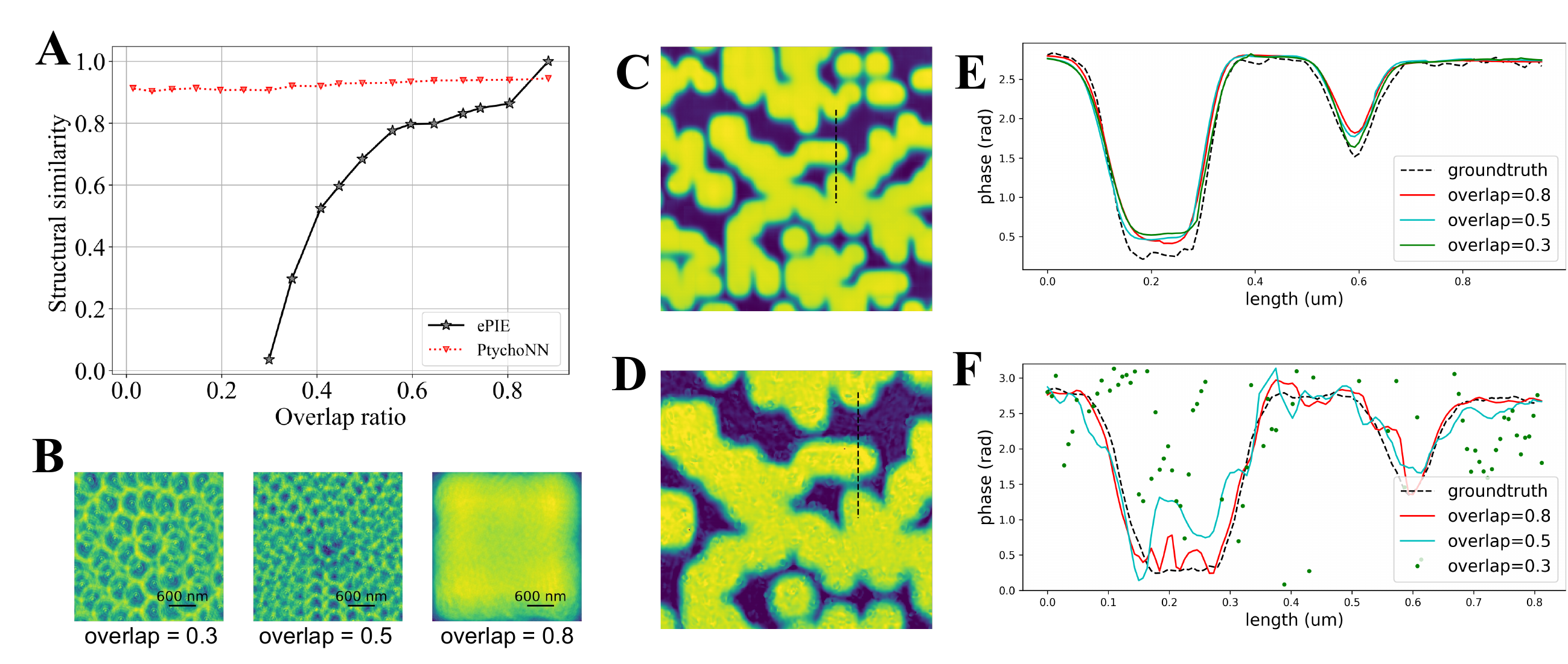}}
\caption{Inference accuracy on sparse sampled data.
\textbf{A} shows the accuracy of both PtychoNN inference and ePIE reconstruction, as a function of the overlap ratio. \textbf{B} shows visualizations of the actual probe overlap for overlap ratios of 0.3, 0.5, and 0.8, by plotting the amplitude of the probe at each scan positions. A defocused probe is used, resulting in a donut-shaped illumination on the sample that can be individually discerned at low overlap. \textbf{C} and \textbf{D} show the phases obtained form PtychoNN and ePIE respectively. \textbf{E} and \textbf{F} show the line profile is extracted at the location of the black dashed line from \textbf{C} and \textbf{D} respectively. For PtychoNN, the inferred phase remains similar to the ground truth regardless of the overlap ratio. For ePIE, the reconstructed phase depends heavily on the overlap ratio and fails completely at overlap = 0.3.}
\label{fig:overlap}
\end{figure}

We then explore the possibility for the workflow to invert sparsely sampled data. Because the inference is performed independently on each diffraction pattern, the notion of spatial overlap no longer applies. An experimental test scan with an overlap ratio of 0.88 is used as the starting point, the ePIE reconstruction of which serves as the ground truth. We then gradually reduce the overlap by selectively removing part of the data. At each step, an ePIE reconstruction is performed alongside stitching of the PtychoNN inference results. The accuracy is evaluated as the structural similarity \cite{ssim} of the reconstructed or inferred phase against the ground truth. As shown in Figure \ref{fig:overlap}\textbf{A}, the accuracy of conventional iterative methods drops rapidly with decreasing overlap ratio, to below 80\% at an overlap ratio of 0.6. In comparison, stitched PtychoNN inference retains over 80\% of accuracy even without any overlap (i.e., when the step size of the scan equals the FWHM of the beam). It can thus be concluded that for an acceptable accuracy of 80\% (90\%), the step size required for the workflow can be 2.5$\times$ (5$\times$) the size required for conventional iterative methods. This in turn indicates that for a given amount of time the workflow can cover 6.25$\times$ (25$\times$) the area measured by conventional ptychographic imaging while simultaneously lowering the dose by the same factor on the sample. The latter is particularly appealing for beam-sensitive samples such as biological materials. 

\subsection*{Robustness in low light}
\begin{figure}[h!]
\centerline
{\includegraphics[width=1\linewidth]{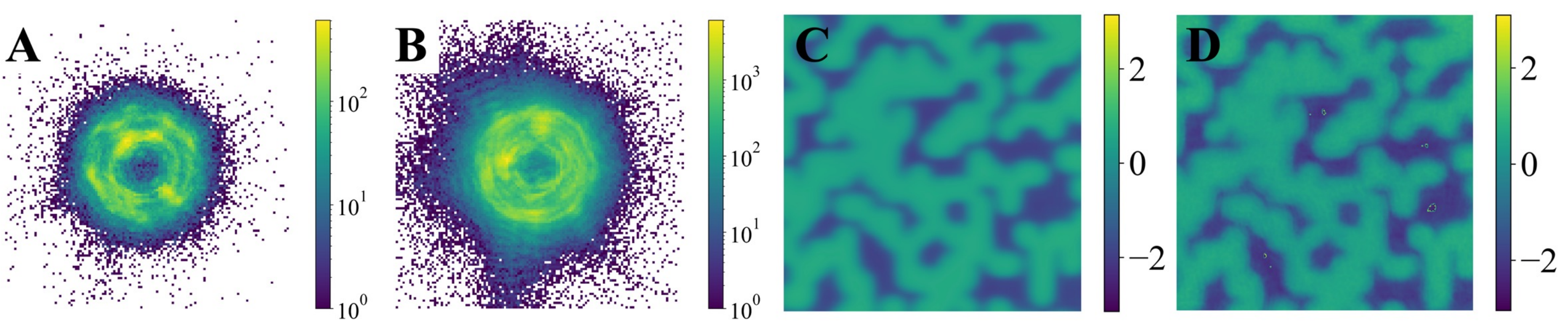}}
\caption{Inference accuracy on low count data. 
\textbf{A} shows an example diffraction pattern of a ptychographic scan with an exposure time of \SI{0.5}{\milli \second}. \textbf{B} shows an example diffraction pattern of the training data with an exposure time of \SI{5}{\milli \second}. \textbf{C} shows the stitched inference results after upscaling the experimental intensity by a factor of 10. \textbf{D} shows the corresponding ePIE reconstructions.}
\label{fig:upscale_int}
\end{figure}

The beam dose can be further reduced by lowering the exposure time of the experimental data. Because the accuracy of the workflow is set by that of the trained model, the count rate of the training data is kept high. We demonstrate two strategies to address the disparity in count rate between training and experimental data. The more straightforward solution is to scale up the experimental intensity before feeding it to the edge device. This approach has the advantage of not requiring retraining of the neural network, but does not work with large scaling factors. For a scaling factor of 100, for instance, a detector pixel with 1 photon count (and a counting uncertainly of 100\%) would have been upscaled to 100 counts (with presumably a counting uncertainly of only 10\%), causing noticeable errors in the inference results. Figure \ref{fig:upscale_int} A shows an example of the experimental data with an exposure time of \SI{0.5}{\milli \second}, corresponding to the maximum detector framerate of \SI{2}{\kilo \hertz}. Figure \ref{fig:upscale_int}B shows an example of the training data, exposed for \SI{5}{\milli \second}. Figure \ref{fig:upscale_int}C shows the stitched inference results after upscaling the experimental intensity by a factor of 10, with an accuracy of 86\% as measured against the groundtruth image reconstructed by the ePIE algorithm (Figure \ref{fig:upscale_int}D). A video recording of the live demonstration of inference at \SI{2}{\kilo \hertz} can be found here\cite{video_ref}. The second solution is to scale down the intensity in the training data and retrain the model. This way a much larger scaling factor can be achieved. As is shown in supplementary Figure \ref{fig:scaled_diff}, an accuracy of over 80\% is observed with a scaling factor of as large as 10,000.

\subsection*{Effect of continual learning}

\begin{figure}[h!]
\centerline
{\includegraphics[width=1\linewidth]{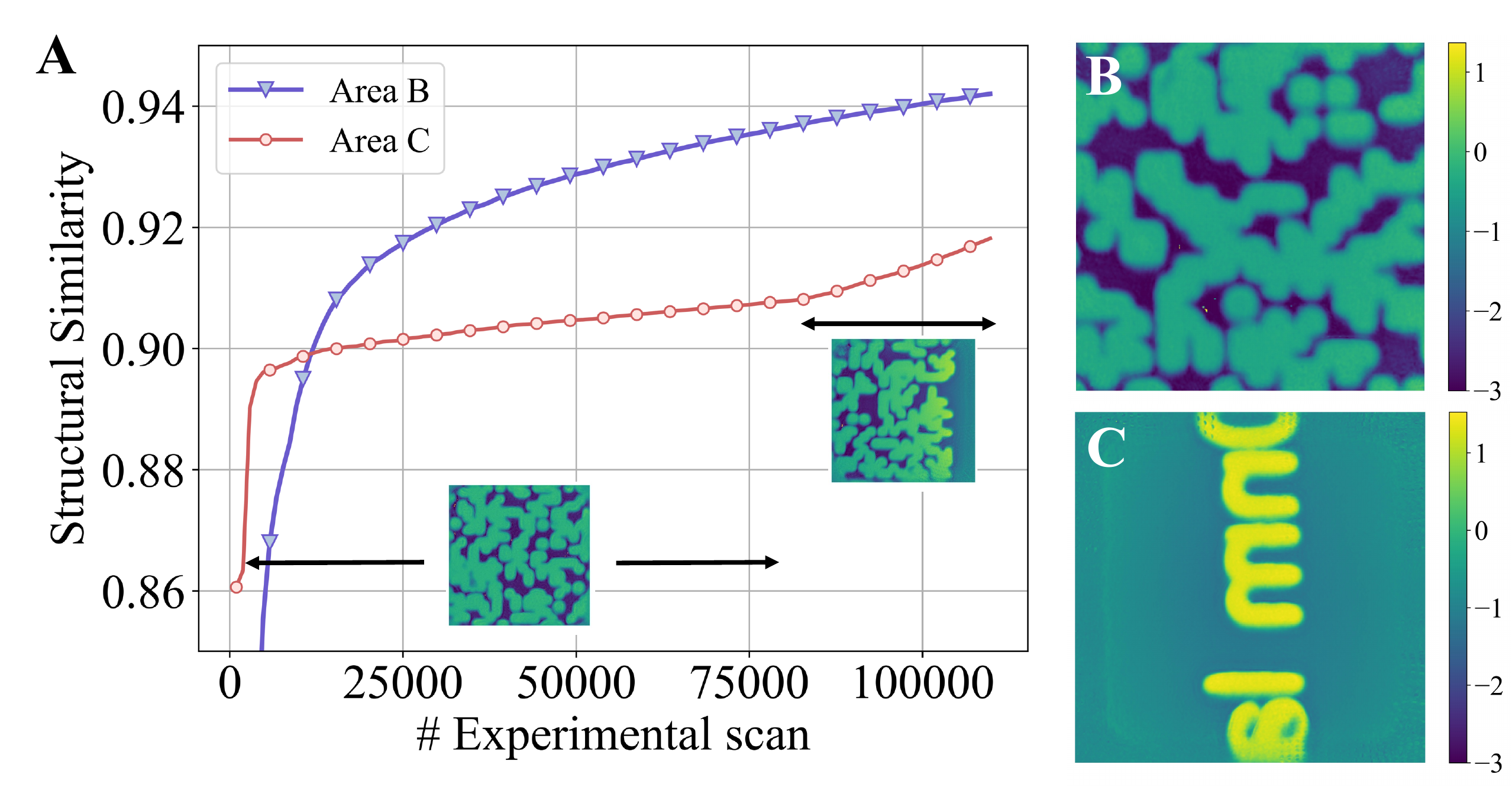}}
\caption{\textbf{A} Evolution of mean squared error (MSE) on two selected test-sample areas during continual learning. Those areas are shown respectively in \textbf{B} and \textbf{C}. The training set consists of pairs of ePIE reconstructed phase and corresponding diffraction patterns on an area with randomly etched features. After about 80,000 sets, the training data starts to include the edge of the patterned areas.}
\label{fig:iteration_error}
\end{figure}

To ensure a high accuracy of the inferred phase, and to quickly adapt to new sample features, the inference model at the edge is constantly updated through continual learning. Figure \ref{fig:iteration_error}A shows the performance of the workflow on two test-sample areas, shown respectively in Figure \ref{fig:iteration_error}B and C. The first area contains features similar to the training data, in which case the mean squared error (MSE) of the training improves rapidly and highly accurate inference is achieved using just the first 20,000 sets of training data. The second area contains feature unseen in the training data. The MSE in this case improves progressively over the course of the continual learning. We note a change of slope in the reduction of MSE after 80,000 sets. This is understood as due to added diversity, as the training data at this point starts to include edges of the patterned areas. The continual learning strategy is thus essential to making accurate inference on new features. The reconstruction fidelity of the neural network over the first few scans is illustrated in supplementary Figure \ref{fig:cont_learning}.

\section*{Discussion}
\begin{figure}[H]
\centering
{\includegraphics[width=0.95\linewidth]{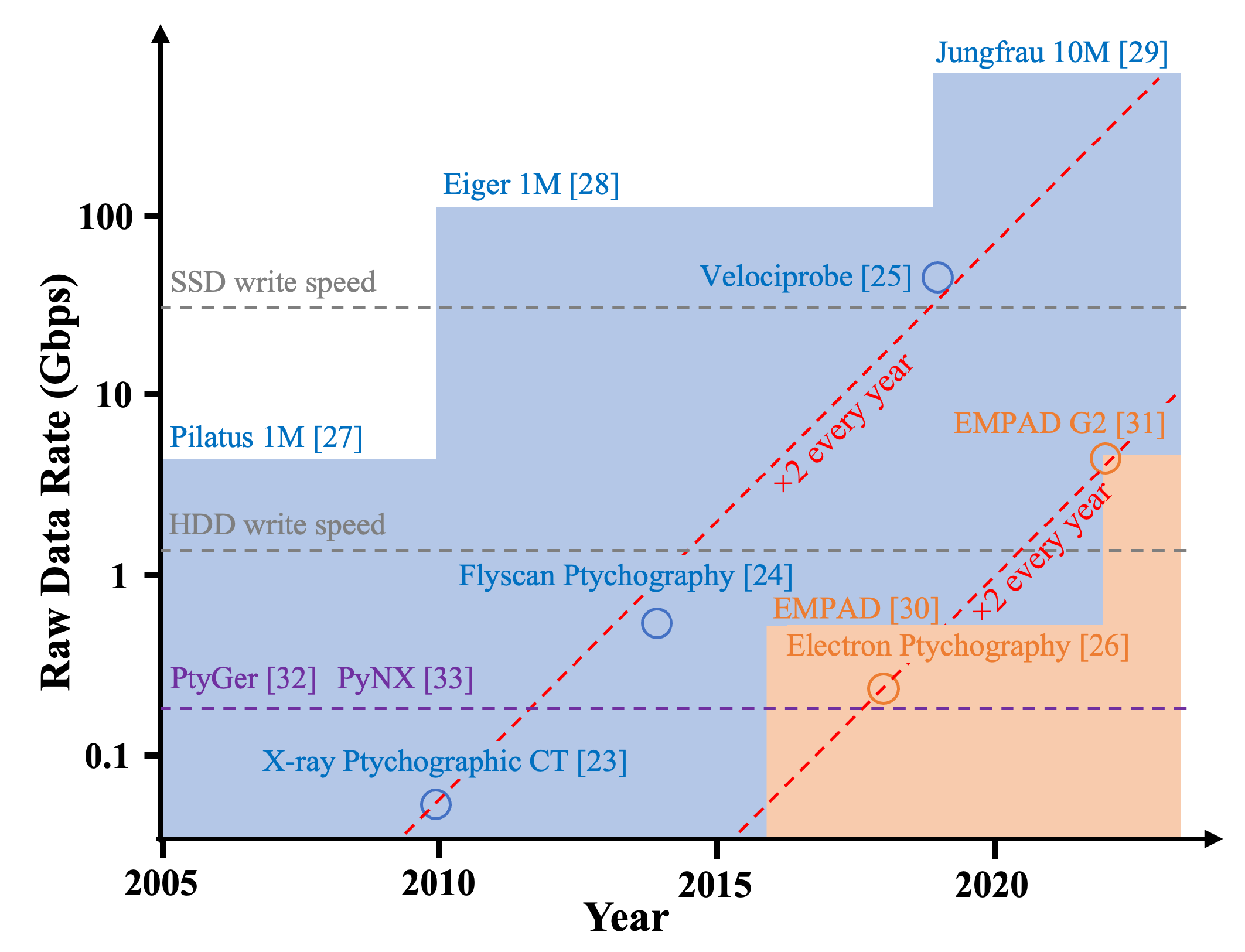}}
\caption{Evolution of raw data rate for state-of-the-art X-ray and electron ptychography.}
\label{fig:fig_data}
\end{figure}

The workflow described in this work provides a path to tackling the data and compute needs for ptychographic imaging at next generation light sources and at advanced electron microscopes. As shown in Figure \ref{fig:fig_data}, the raw data rate for both state-of-the-art X-ray \cite{ptycho_tomo, flyscan_ptycho, velociprobe} and electron \cite{second_e_ptycho} ptychography doubles every year, enabled by the development of both advanced scanning strategies and larger and faster detectors \cite{Pilatus_1M, Eiger_1M, jungfrau, EMPAD_2016, EMPAD-G2}. For X-ray ptychography, the raw data rate reached \SI{800}{Mbps} a decade ago, and this still exceeds the throughput of the most advanced iterative optimizers (in 2021) \cite{Ptyger,PyNX}. To put this into perspective, it would take conventional methods an hour to perform phase retrieval on data taken in a second on a fully illuminated 10 Mega-pixel detector running at 32 bits and \SI{2}{\kilo \hertz} (\SI{640}{Gbps}). Our workflow overcomes this by leveraging the high performance of surrogate deep learning models, although the inference speed in our demonstration is ultimately limited by the \SI{1}{Gbps} network connection on the detector control computer. For a detector image size of 512$\times$512 pixels, live inference at \SI{100}{\hertz} is achieved, corresponding to a capped incoming data rate of \SI{0.5}{Gbps}. By reducing the image size to 128$\times$128 pixels, live inference at \SI{2}{\kilo \hertz} is achieved while running the detector at its maximum frame-rate. 

The accuracy of the workflow improves over the course of the experiment via continual learning and given sufficient data, is ultimately limited only by the accuracy of the iterative phase retrieval used for the training. Because access to on-demand HPC resources may be limited, a strategy is devised to minimize use of expensive HPC resources. In the initial phase, when a new set of experimental data is received at the HPC, it automatically triggers the image reconstruction service using iterative methods. The results are then cropped and paired with the corresponding diffraction patterns to form the labeled training data. Part of this training data is used immediately to validate the accuracy of the existing neural network. If a large mismatch is detected between the inferred phase from the latest model and the iteratively retrieved phase in the training data, a retraining service is queued which upon completion also sends the updated model to the edge device. If the mismatch is within a certain tolerance level (for instance, a difference of less than 10\% in structural similarity), the automatic image reconstruction and retraining services are suspended, with the experimental data being sent and reconstructed at HPC only at a prolonged interval (for instance, once every hour). The on-demand image reconstruction and retraining services are resumed if a large mismatch is again detected. Typically, a new network is trained from scratch at the beginning of each experiment to better accommodate for differences in samples, probe energy and detector distance. However, more sophisticated schemes such as the one described in the fairDMS approach\cite{fairDMS} can be adopted to reuse models from previous experiments for achieving more rapid training of the data.

One significant advantage of the proposed workflow is in the field of low-dose high resolution imaging, which is relevant for a variety of materials including biological samples and organic-inorganic perovskites. Conventional ptychography is in particular damaging to these materials due to the overlapping constraint. We note that beam damage is an even bigger problem for electron ptychography than for X-ray due to the stronger interaction with materials of the electron beam. In this work, low-dose high resolution imaging is achieved through two separate approaches. First, by inferring on individual diffraction patterns, the oversampling constraint is eliminated. We show sample image with an accuracy of 90\% without any probe overlap between the measured points. This reduces the beam dose by a factor of 25 compared to that required by conventional methods. In the second approach, the beam dose is further reduced by a factor of 10 to 10,000 by inferring on low counts data. We show that it is possible to retain over 85\% accuracy by either downscaling the training data taken at higher count rates or upscaling the low-count rate experimental data. The former approach has the advantage of working with extremely low-count data, while the latter is convenient as it does not require retraining of the existing model. Finally, we note that both approaches, in addition to reducing beam damage by a combined factor of about 1,000, increase the measurement area per given time by the same amount.

\section*{Methods}

\subsection*{Experimental Details for Ptychography}

\color{black}
X-ray ptychography data was taken at the hard X-ray nanoprobe beamline at the Advanced Photon Source. The photon energy was 10 keV. Two sets of experimental data was taken. The first set was taken with an Amsterdam Scientific Instruments Medipix3  detector (516$\times$516 pixels, \SI{55}{\micro \meter} pixel size) sitting at \SI{1.55}{\meter} downstream of the sample. The maximum frame rate was \SI{100}{\hertz} limited by the data transfer bandwidth of the \SI{1}{Gbps} network. The second set was taken with a Dectris Eiger2 X 500K detector with \SI{75}{\micro \meter} pixels, located \SI{0.9}{\meter} downstream from the sample. Only a subsection of the 128$\times$128 image is used, allowing live inference at the maximum detector frame rate of \SI{2}{\kHz}.

\subsection*{Neural Network Architecture and Training}

At the heart of the workflow is a convolutional neural network that takes as input a raw coherent diffraction image and outputs an inferred sample structure in a single pass (live inference). A modified version of PtychoNN \cite{PtychoNN} was used for the online inference. Supplementary Figure \ref{fig:SF-model_comp} shows the qualitative comparison of the inferences made by modified PtychoNN (PtychoNN 2.0) against the original network. Unlike previous implementations of PtychoNN and neural networks doing phase retrieval \cite{PtychoNet, DNN_ptycho1, DNN_ptycho2, DNN_ptycho3}, training is performed online in this workflow, i.e., it is updated repeatedly as new diffraction data are obtained from the experiment. Each time a new batch of data is acquired, the data are reconstructed to generate phase-retrieved images, the diffraction data plus the phase-retrieved images are appended to the existing corpus of training data, and PtychoNN is trained for a further 50 epochs by using a cyclic learning rate policy~\cite{smith2017cyclical}. Network weights are updated by using the adaptive moment estimation (ADAM) optimizer to minimize the mean absolute error (MAE) between the target labels (ePIE output) and  network inferences. The model state at each training epoch is evaluated on unseen validation data, which is 10\% of the training data. The trained model with the lowest validation loss at the end of 50 epochs is pushed to the edge device for inference on subsequent scans. A total of 113 experimental scans, each with 963 X-ray diffraction patterns were acquired. The total available training data thus comprised 113$\times$963 = 108,819 pairs of diffraction and sample images. 

\subsection*{Real-time phase retrieval at higher frame rates}
The embedded GPU device (NVIDIA Jetson) shown in Fig. \ref{fig:overall-flow} can support real-time feedback up to a frame rate of \SI{100}{\hertz}. Data acquisition rates for ptychography experiments are growing rapidly and are soon expected to exceed \SI{50}{Gbps}. Therefore, it is important to reduce inference times to achieve real-time data analysis. Widely studied and adopted techniques like reduced precision arithmetic and quantization can be used to accelerate inference \cite{quant1}. However, such methods can degrade network accuracy and require careful fine tuning of hyper parameters. Since the neural network in study, PtychoNN, is a fully convolutional neural network, and is compute-limited, we explored the use of an advanced GPU that is designed to accelerate ML workloads. 
We benchmarked the inferences times of a Ampere-architecture GPU (RTX A6000) for high-speed inference with different batch sizes as shown in Table \ref{table:perf_comp2}. The Dectris Eiger2 X 500K detector used for high-speed data acquisition can support a maximum data rate of \SI{2}{\kilo \hertz} at 16 bits. A batch size of 8 was chosen for inference at \SI{2}{\kilo \hertz} as the inference time per frame is observed not to improve substantially for higher batch sizes. Table \ref{table:perf_comp2} shows the average inference times, plus standard deviations, as measured across 50 inference runs.  

\begin{table}[h!]
	\caption{Approximate inference times (\SI{}{\micro \second}) per frame on the RTX A6000 \& AGX Xavier}
	\label{table:perf_comp2}
	\centering 
		\begin{tabular}{|c|c|c|}
			\hline
			Batch size \# & RTX A6000 & AGX Xavier \\
			\hline
			1  &  370 $\pm$ 20 &  2300 $\pm$ 400 \\
			\hline
			2   & 220 $\pm$ 20 &  1360 $\pm$ 340\\
			\hline 
			4   & 130 $\pm$ 10 &  960 $\pm$ 110 \\
			\hline 
			8   & 90 $\pm$ 5 &  850 $\pm$ 140\\
			\hline 
			16   & 70 $\pm$ 5 &  680 $\pm$ 30 \\
			\hline 
		\end{tabular}
\end{table}

\section*{Acknowledgements}
This research used resources of the Advanced Photon Source (APS), Center for Nanoscale Materials (CNM), and Argonne Leadership Computing Facility (ALCF), which are operated for the DOE Office of Science by Argonne National Laboratory under Contract No. DE-AC02-06CH11357. This work was also supported by the U.S. Department of Energy, Office of Science, Office of Basic Energy Sciences Data, Artificial Intelligence and Machine Learning at DOE Scientific User Facilities program under Award Number 34532. M.J.C.\ and S.K.\ also acknowledge support from Argonne LDRD 2021-0090 – AutoPtycho: Autonomous, Sparse-sampled Ptychographic Imaging. We gratefully acknowledge the computing resources provided on Swing, a high-performance computing cluster operated by the Laboratory Computing Resource Center at Argonne National Laboratory. We gratefully acknowledge insightful discussion and advice from Francesco De Carlo at the Advanced Photon Source. We would also like to thank William Allcock and his team at the ALCF for their help with HPC resources used in this work, including the on-demand computing testbed. 

\subsection*{Data and materials availability:}
The code for the inference implemented at the edge is available at  \url{https://github.com/vbanakha/edgePtychoNN} and the data will be made available at \url{https://anl.app.box.com/folder/157836921891}. All other data needed to evaluate the conclusions in the paper are present in the paper and/or the Supplementary Materials.

\renewcommand{\figurename}{Supplementary Figure}
\renewcommand\thefigure{\arabic{figure}}
\setcounter{figure}{0}  

\section*{Supplementary materials}

\begin{figure}[h!]
\centerline
{\includegraphics[width=0.7\linewidth]{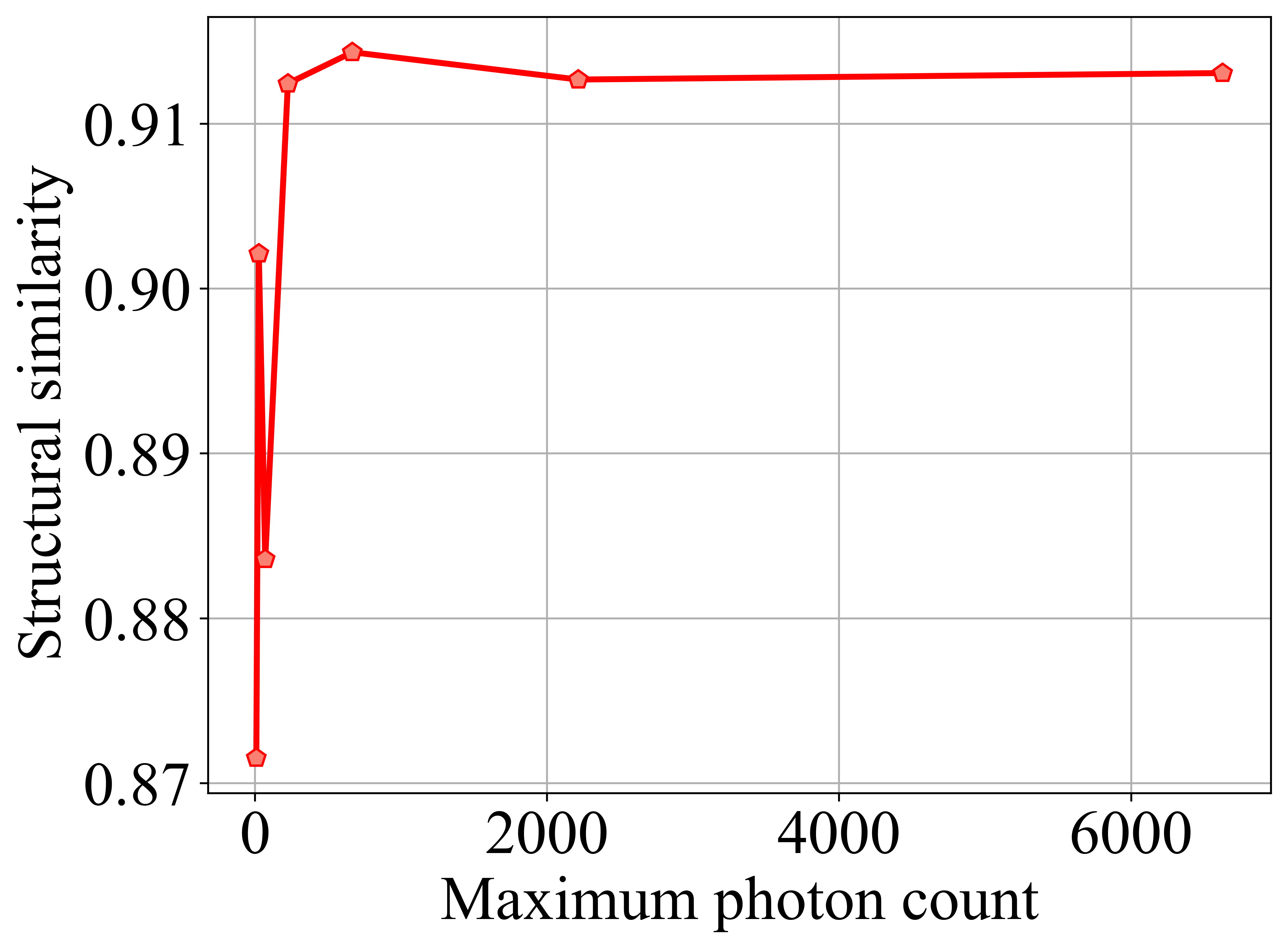}}
\caption{Performance of PtychoNN when the diffraction intensities are logarithmically scaled down with scaling factors starting from 10 to 10,000. The x-axis represents the maximum photon count per diffraction pattern at each of the scaling factors. It can be observed that the neural network can retain a structural similarity above 80\% when the maximum photon count per diffraction pattern is as low as 10.}
\label{fig:scaled_diff}
\end{figure}

\begin{figure}[h!]
\centerline
{\includegraphics[width=0.5\linewidth]{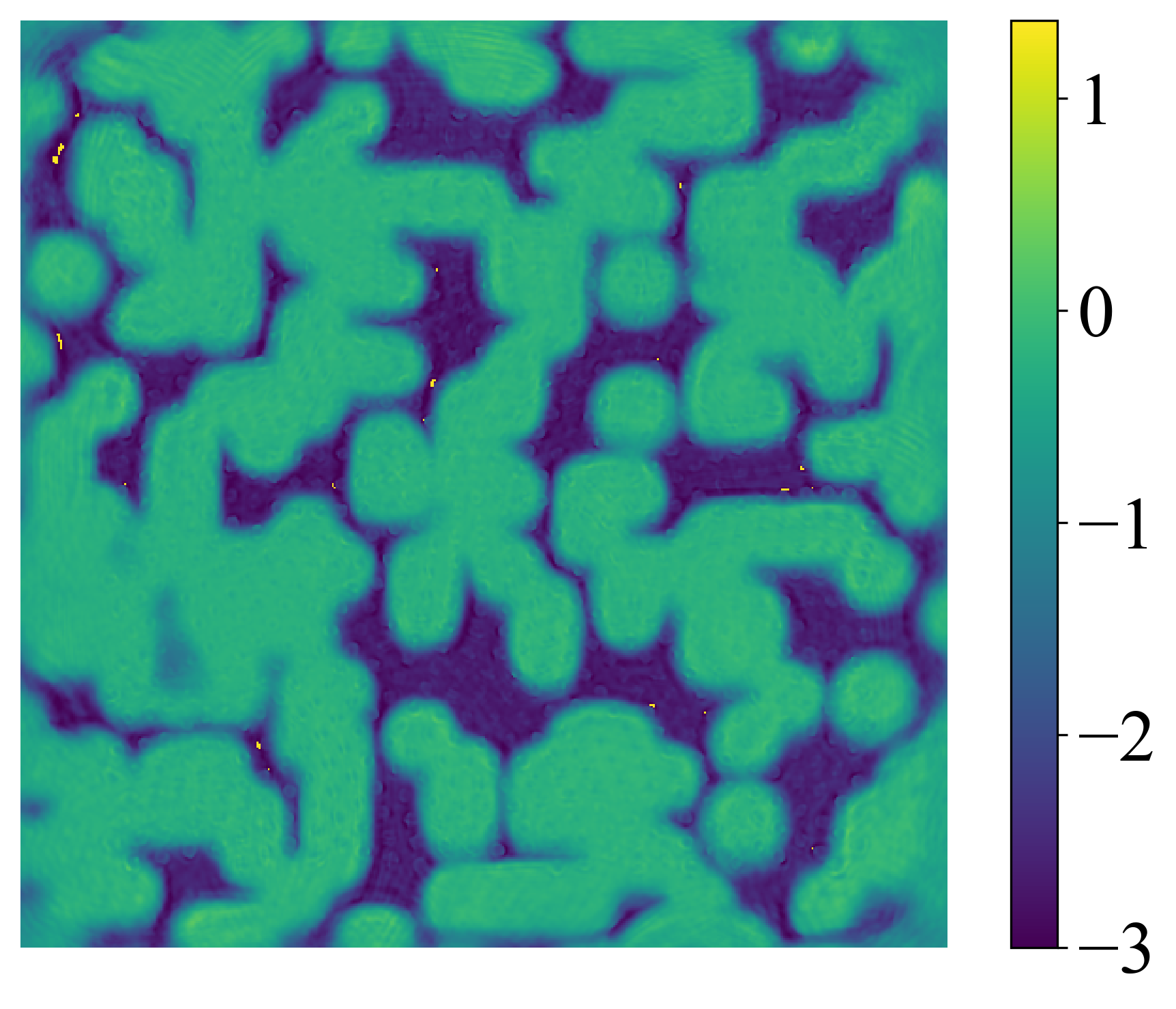}}
\caption{An example of ePIE reconstructed experimental training data included in the online training of the neural network.}
\label{fig:SF_trainingset}
\end{figure}

\begin{figure}[h!]
\centerline
{\includegraphics[width=0.7\linewidth]{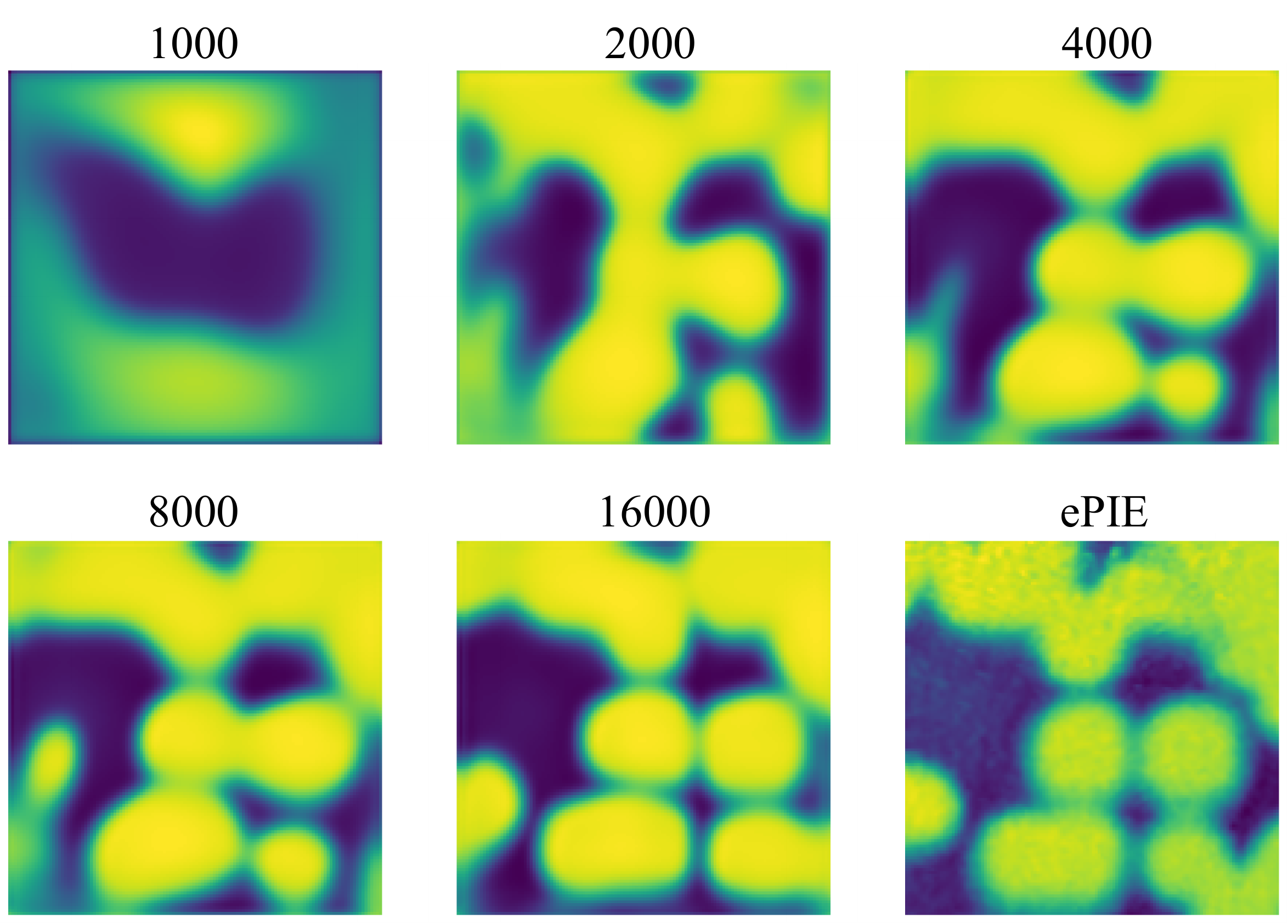}}
\caption{Evolution of neural network predictions as a function of experimental training data compared to the iterative ePIE reconstruction.}
\label{fig:cont_learning}
\end{figure}

\color{black}
\subsection*{NVIDIA's Jetson AGX Xavier Developer Kit}

We used the NVIDIA Jetson AGX Xavier developer kit \cite{AGX}, an embedded GPU platform, to demonstrate real-time ptychographic phase retrieval at the edge. AGX Xavier is one among the Jetson series from NVIDIA with a computational capability of 32 TOPs, dedicated for building embedded ML solutions \cite{AGX_specs}. Here the AGX Xavier is targeted as an inference device and the TensorRT$^{TM}$ Python API from NVIDIA is used for accelerating the inference workflow. One of the primary means of importing the model in TensorRT$^{TM}$ is via the ONNX format, and therefore the PyTorch trained model is converted to ONNX before running the inference workflow. 

\subsection*{PtychoNN 2.0}
\begin{figure}[h!]
\centerline
{\includegraphics[width=1\linewidth]{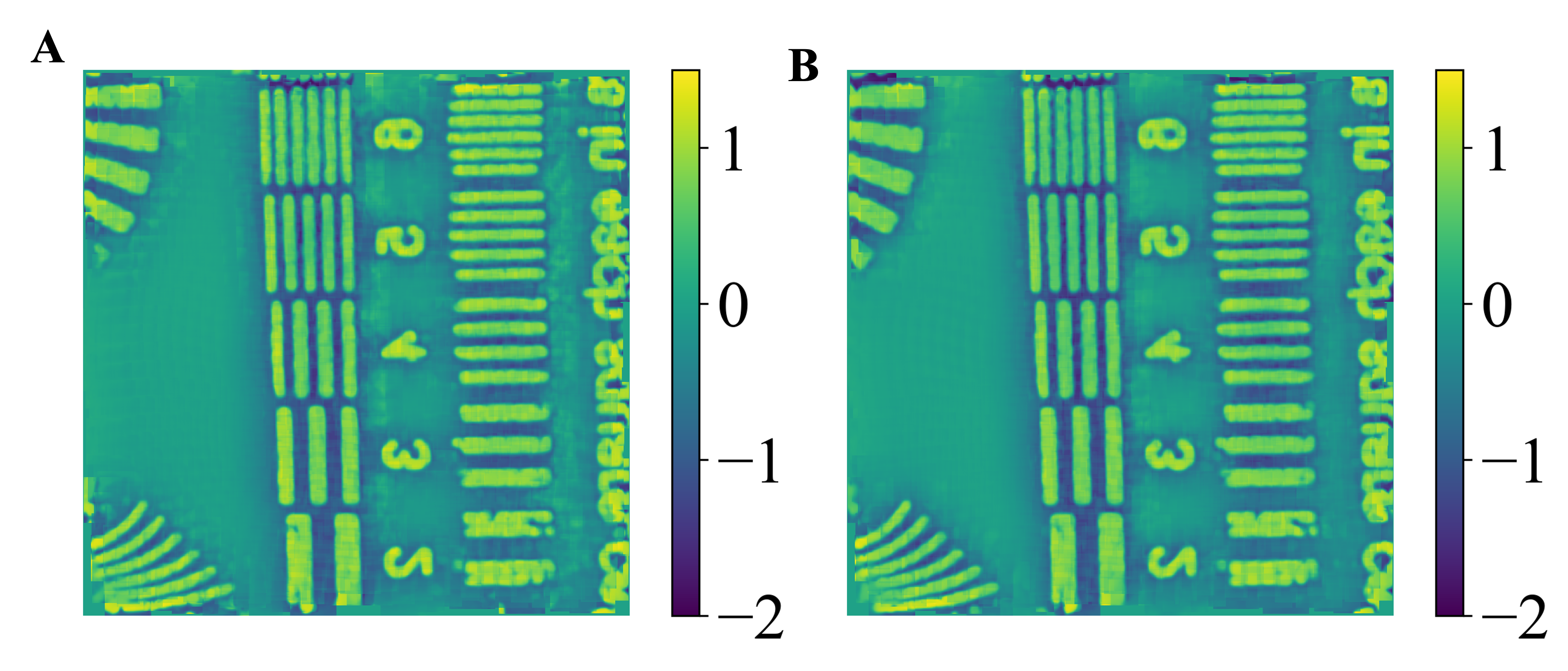}}
\caption{Qualitative comparison of inferences by two different models of PtychoNN, \textbf{A} having 4.8 $\,$M trainable parameters, and \textbf{B} PtychoNN 2.0 with 0.7 $\,$M trainable parameters.}
\label{fig:SF-model_comp}
\end{figure}

As the edge device also supports native PyTorch implementations, we compared the inference times for PyTorch against TensorRT for the ML models in study. We considered two ML models, one is the fully convolutional network discussed in \cite{PtychoNN} with 4.7$\,$M parameters and the other is a slightly modified network architecture (PtychoNN 2.0) with 0.7$\,$M parameters. Supplementary Figure \ref{fig:SF-model_comp} indicates that PtychoNN 2.0 can give equally good predictions as PtychoNN. Table \ref{table:perf_comp1} shows the average execution times observed on AGX, over 50 iterations for each of the models in TensorRT and PyTorch for a batch size of $1$. The AGX was operated in Max-N mode during these inference runs. It has to be noted that the inference times shown in table excludes the pre-processing time. PtychoNN in table \ref{table:perf_comp1} corresponds to the network discussed in \cite{PtychoNN} and PtychoNN 2.0 refers to the light weight model designed for faster ptychographic reconstructions. TensorRT is observed to exhibit a speed up of approximately 4$\times$ when compared to native PyTorch implementation for PtychoNN 2.0. 
\begin{table}[h!]
	\caption{Approximate inference times (\SI{}{\milli \second}) in TensorRT and PyTorch on the AGX}
	\label{table:perf_comp1}
	\centering 
		\begin{tabular}{|c|c|c|}
			\hline
			Model \# & TensorRT & PyTorch \\
			\hline
			PtychoNN  & 10 $\pm$ 1 & 15 $\pm$1  \\
			\hline
			PtychoNN 2.0   & 2.3$\pm$0.4  & 8 $\pm$1 \\
			\hline 
			
		\end{tabular}
\end{table}

\medskip

\bibliographystyle{Science}
\bibliography{scibib}

\end{document}